\documentclass[journal,twoside,web]{ieeecolor}

\usepackage{generic}
\usepackage{cite}
\usepackage{amsmath,amssymb,amsfonts}
\usepackage{graphicx}
\usepackage{textcomp}

\makeatletter
\def\thebibliography#1{%
  \section*{References}%
  \list{\@biblabel{\arabic{enumiv}}}%
  {\settowidth\labelwidth{\@biblabel{#1}}%
   \leftmargin\labelwidth
   \advance\leftmargin\labelsep
   \usecounter{enumiv}}%
  \sloppy
  \clubpenalty4000
  \widowpenalty4000
  \sfcode`\.=\@m}

\makeatother

\begin{document}

\title{3D Geometric Tooth Alignment Planning via Deep Reinforcement Learning}

\author{Yong Li, Jianwen Lou, Jiayue Ma, Yao-Xiang Ding, Youyi Zheng, and Haihua Zhu
\thanks{This work was supported in part by Choho Technology, which provided the dataset used in our experiments.}
\thanks{Yong Li and Jianwen Lou are with the School of Software Technology, Zhejiang University, Hangzhou, China (e-mail: 22451201@zju.edu.cn, jianwen.lou@zju.edu.cn). }
\thanks{Jiayue Ma, Yao-Xiang Ding, and Youyi Zheng are with the State Key Laboratory of CAD\&CG, Zhejiang University, Hangzhou, China.}
\thanks{Haihua Zhu is with the Stomatology Hospital, Zhejiang University School of Medicine, Hangzhou, China.}
\thanks{Corresponding author: Jianwen Lou.}}

\maketitle

\begin{abstract}
3D geometric tooth alignment planning, which determines sequential trajectories from initial malocclusion to the final target alignment, is a cornerstone of modern digital orthodontics. This paper presents a novel deep reinforcement learning (DRL) framework to automate the generation of these alignment paths. We formulate the planning process as a Markov Decision Process (MDP) to capture its sequential decision-making nature, focusing on optimizing geometric trajectories while integrating essential spatial constraints, such as inter-dental collision avoidance and path efficiency. The proposed method leverages the Deep Deterministic Policy Gradient (DDPG) algorithm, enhanced by three key innovations: (1) a Transformer-based agent to model complex spatial interactions between teeth and manage high-dimensional state-action spaces; (2) a dynamic masking scheme that restricts movement to a sparse subset of teeth per step, better reflecting the clinical logic of sequential alignment; and (3) a two-stage curriculum learning strategy that gradually increases task difficulty to ensure training stability and efficient path discovery. We evaluate our approach on a dataset of 10K expert-designed treatment plans based on clinical data. Experimental results demonstrate that our method outperforms existing baselines in terms of path safety and geometric efficiency, providing a robust and automated solution for 3D geometric orthodontic alignment planning.
\end{abstract}


\section{Introduction}
\label{sec:introduction}
3D geometric tooth alignment planning is a fundamental component of digital treatment simulations in orthodontics. Utilizing high-resolution 3D intra-oral scans as the primary data modality, this process is responsible for generating sequential motion trajectories that transition teeth from initial malocclusion to the target alignment. As a cornerstone of modern digital orthodontics~\cite{wu2022twostage, liu2023hierarchical}, this process ensures collision-free, spatially efficient, and personalized geometric transitions. Despite its critical role in automated treatment workflows, this planning process is primarily performed manually by practitioners in current clinical practice~\cite{wang20243d}, making it a time-consuming and laborious task. As a result, there is a strong need for a fully automated solution. However, the complexity of real-world dentition, such as missing teeth, crowding, and varying tooth morphologies, along with the necessity to satisfy intricate spatial constraints, makes automated 3D geometric tooth alignment planning a challenging problem.

Traditional approaches to this task have framed the problem as a constrained optimization task, using algorithms like Particle Swarm Optimization (PSO)~\cite{xu2020orthodontic, ma2021orthodontic} and Gray Wolf Optimization (IGWO)~\cite{du2023orthodontic} to identify the optimal tooth movement path. However, these methods are limited by their simplistic frameworks. They typically focus on basic criteria, such as path smoothness and minimizing displacement, but fail to account for more complex factors like intricate tooth-to-tooth spatial relationships, which are crucial for accurate planning. As a result, the generated movement paths often do not meet clinical standards. Additionally, these approaches are case-specific, restricting their adaptability. In response to these limitations, recent research has turned to using deep neural networks, such as Transformers~\cite{ma2024neural} and Diffusion Models~\cite{fan2024collaborative, fan2026progressive}, to learn intricate tooth movement patterns directly from expert-designed data. These data-driven methods offer greater efficiency and produce results that align more closely with real-world practices. However, current deep learning approaches often treat the task as a sequence-to-sequence generation problem, where the entire trajectory is generated at once with a predefined length. This overlooks the sequential decision-making nature of clinical alignment, where tooth movements should be predicted step-by-step based on both the current and target dentition.

This study proposes a novel deep reinforcement learning-based approach to automated 3D geometric tooth alignment planning. We frame the process as a Markov Decision Process (MDP) to capture its sequential decision-making nature, incorporating essential constraints such as safety (e.g., avoiding inter-dental collisions) and efficiency (e.g., minimizing trajectory length). The proposed method utilizes a Deep Deterministic Policy Gradient (DDPG) algorithm~\cite{lillicrap2015continuous}, augmented with three key adaptations: (1) a transformer-based agent to model complex tooth interactions and manage high-dimensional state and action spaces, (2) a dynamic masking scheme to ensure only a sparse set of teeth are moved at each step, mimicking clinical logic, and (3) a two-stage curriculum learning approach to progressively impose stricter conditions on the training process, reducing exploration challenges and improving training stability. We validate the proposed method on a dataset containing 10K expert-designed orthodontic pathways based on clinical data. The results demonstrate that our method generates safer and more efficient tooth movement paths compared to existing methods, establishing a new state-of-the-art.

In summary, the main contributions of this work are:
\begin{itemize}
\item We introduce the first deep reinforcement learning-based approach for 3D geometric tooth alignment planning, modeling the problem as an MDP that fully captures its sequential decision-making nature.
\item We adapt the DDPG algorithm with three novel modifications, creating an effective framework to handle dental intricacies and high-dimensional state-action spaces.
\item We conduct extensive experiments, demonstrating the effectiveness of the proposed method and providing deep insights into its key components.
\end{itemize}

\section{Related Work}
\label{sec:related work}
Automated 3D geometric tooth alignment planning is being explored through two primary approaches: optimization-based methods and deep learning-based methods.
\subsection{Optimization-based Methods}
Early works frame 3D geometric tooth alignment planning as a constrained optimization challenge, using algorithms such as Artificial Bee Colony (ABC)~\cite{li2020orthodontic}, Particle Swarm Optimization (PSO)~\cite{xu2020orthodontic, ma2021orthodontic}, and Gray Wolf Optimization (IGWO)~\cite{du2023orthodontic} to search for optimal movement paths. However, these approaches are limited by overly simplified optimization frameworks. Specifically, the path length is often predetermined, and the constraints remain basic, primarily focusing on trajectory smoothness and minimizing total displacement. Crucial geometric features, such as the intricate morphological interactions between teeth, are difficult to incorporate into these frameworks. As a result, the generated trajectories often fail to achieve collision-free and clinically viable alignment. Additionally, these methods are case-specific, optimizing each path individually without the ability to learn from or generalize across large-scale historical datasets. This significantly limits their adaptability to real-world scenarios involving diverse and complex dentitions.

\subsection{Deep Learning-based Methods}
The limitations of optimization-based approaches have driven a paradigm shift toward deep learning-based methods. Recent studies leverage deep neural networks, such as Transformers~\cite{ma2024neural} and Diffusion Models~\cite{fan2024collaborative}, to fit expert-designed alignment paths. By utilizing neural networks, these methods can learn intricate movement patterns and spatial features from the training data. Compared to optimization-based methods, they are more efficient and produce results that better align with real-world clinical cases. However, these methods typically treat the task as a sequence-to-sequence generation problem, where the desired path length is predefined and the trajectory is generated in a single pass. This paradigm overlooks the sequential decision-making nature of the process, which in practice involves determining tooth motion step-by-step by evaluating the current dentition relative to the target alignment. 

In recent years, deep reinforcement learning (DRL) has garnered significant attention as a powerful method for solving complex, sequential decision-making problems, including path planning. DRL combines the flexibility of reinforcement learning (RL) with the representational power of deep neural networks, enabling agents to learn optimal policies through trial and error. For example, in robotics, DRL has been extensively applied to constrained motion planning, where deep Actor-Critic methods like DDPG~\cite{lillicrap2015continuous} allow high-dimensional systems to find collision-free paths for complex tasks. While this extensive body of research~\cite{sutton1991dyna,mirowski2016learning, zhu2017target} demonstrates DRL's effectiveness for general path planning, its application to the specialized, spatially constrained domain of 3D geometric tooth alignment planning remains largely unexplored. This study aims to bridge this critical gap.

\begin{figure*}[t]
    \centering
    \includegraphics[width=0.9\textwidth]{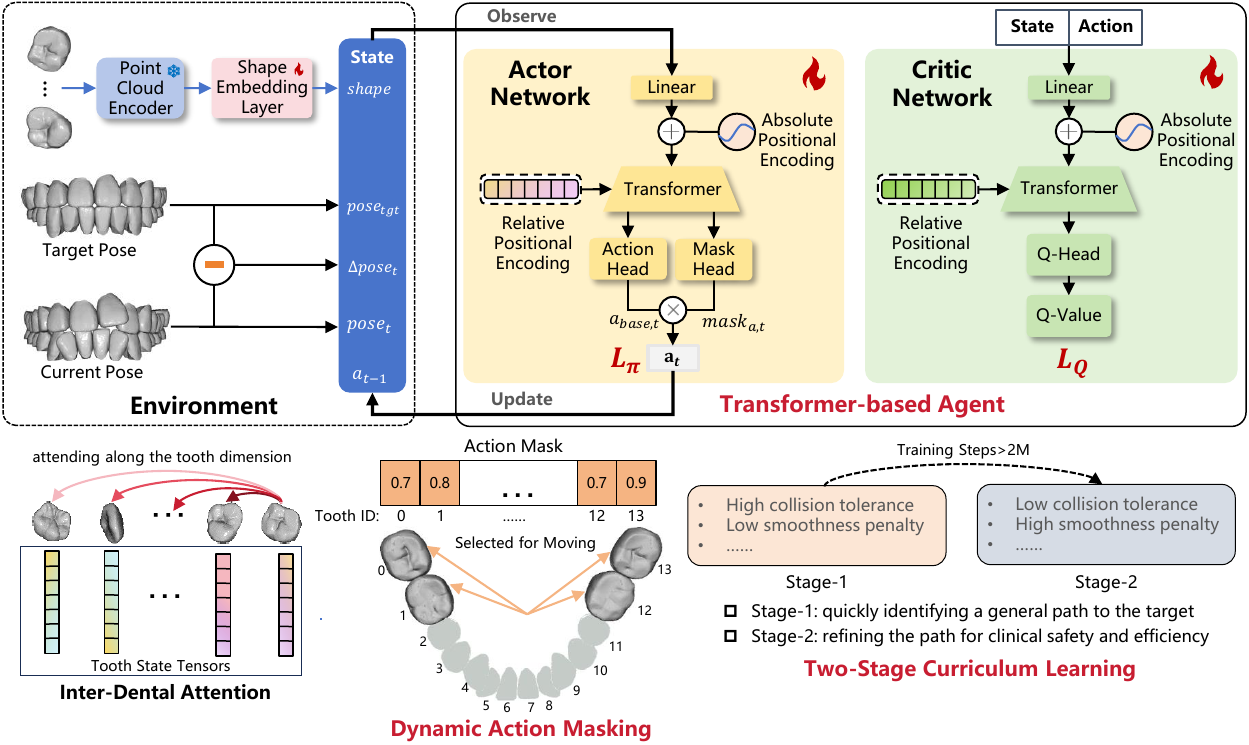}
    \caption{An overview of the proposed method. The method uses a transformer-based agent, consisting of an actor and a critic network, to observe dental states and produce actions for tooth movement. It employs attention mechanisms to model inter-tooth relationships. The method also includes two key components: 1) dynamic action masking for sparse actions, mimicking clinical practice where only a few teeth are moved per step; 2) a two-stage curriculum learning strategy to enhance model training.}
    \label{fig:pipeline}
\end{figure*}

\section{Methodology}
\label{sec:methodology}

\subsection{Overview}
3D geometric tooth alignment planning schedules intermediate tooth configurations, transitioning from an initial malocclusion to the target alignment. This task is inherently a sequential decision-making problem under complex spatial constraints, where the generated trajectories must satisfy both safety (e.g., ensuring collision-free movements) and efficiency (e.g., maintaining smooth and optimal paths). Furthermore, the planning process must account for intricate tooth interactions and irregular arrangements, such as those caused by missing teeth. To address these challenges, we propose a novel deep reinforcement learning-based framework that provides a fully automated, data-driven solution. Our approach begins by framing the alignment planning problem as a Markov Decision Process (MDP) to capture its sequential nature. The key MDP components—state space, action space, and reward function—are meticulously defined within the context of geometric orthodontics. We then employ a customized Deep Deterministic Policy Gradient (DDPG)~\cite{lillicrap2015continuous} algorithm (see Fig.~\ref{fig:pipeline}) to learn an optimal policy, enhanced by three key adaptations: 

\begin{itemize}
\item Transformer-based Agent: The actor and critic networks leverage a transformer architecture to model complex spatial interactions between teeth and manage the high-dimensional state-action spaces inherent in multi-tooth coordination.
\item Dynamic Masking Scheme for Action Sparsity: This scheme ensures the agent moves only a sparse subset of teeth at each step, accurately reflecting the clinical logic of sequential alignment while reducing the complexity of the action space.
\item Two-stage Curriculum Learning: The training process starts with relaxed spatial constraints and gradually introduces stricter conditions, effectively mitigating exploration challenges and ensuring stable policy convergence.
\end{itemize}

\subsection{3D Geometric Tooth Alignment Planning as an MDP}
3D geometric tooth alignment planning involves determining a sequence of tooth movements to achieve collision-free and efficient alignment. Each step in the sequence is predicted by analyzing the current dentition relative to the target alignment, ensuring that each incremental movement transitions the teeth toward the final goal while satisfying spatial constraints. This process can be mathematically modeled as a Markov Decision Process (MDP), where an agent interacts with the geometric environment over discrete steps, making a series of sequential decisions. The goal is to learn an optimal policy that maximizes the expected cumulative reward, representing the most efficient and safe path. In this context, we define the key components of the MDP as follows:

\subsubsection{State Space ($\mathcal{S}$)} 
$s_t \in \mathcal{S}$ represents a snapshot of the environment at timestep $t$. In this study, the environment consists of the current and target dentition, where the pose and shape information of the teeth are the key features. We use the following five components to fully capture these features:
\begin{itemize}    
\item Current Tooth Pose - $pose_{i,t}$: This includes the current position $p_{i,t} \in \mathbb{R}^3$ and orientation $\theta_{i,t} \in \mathbb{R}^6$ of a tooth (where $i$ is the tooth index and $t$ is the timestep).    
\item Target Tooth Pose - $pose_{i,tgt}$: The desired final position and orientation of a tooth.    
\item Current-to-Target Pose Difference - $\Delta{pose}_{i,t}$: The deviation of the current tooth pose $pose_{i,t}$ from the target pose $pose_{i,tgt}$. This information provides an explicit error signal to guide the learning of the policy.    
\item Last Action - $a_{i,t-1}$: The agent's action taken on a tooth in the previous timestep, which consists of a translation $\Delta{p}_{i,t-1} \in \mathbb{R}^3$ and rotation $\Delta{\theta}_{i,t-1} \in \mathbb{R}^6$. This helps the policy consider past actions, thereby encouraging temporal smoothness in the generated path.    
\item Tooth Shape Embedding - $shape_{i}$: A 108-dimensional feature vector that encodes essential shape information of a tooth. This embedding is extracted using a 3D point cloud encoder, which is pretrained via an encoder-decoder framework for a 3D point cloud completion task~\cite{yu2021pointr}.
\end{itemize}

Consequently, the state of a tooth can be represented as $[pose_{i,t},pose_{i,tgt},\Delta{pose}_{i,t},a_{i,t-1},shape_{i}] \in \mathbb{R}^{144}$. Thus, for the state $s_t$ of 28 teeth\footnote{Each 3D dental model in this study includes both lower and upper jaw teeth, resulting in a total of up to 28 teeth (excluding wisdom teeth, as in previous studies \cite{ma2024neural, fan2024collaborative})}, the representation becomes a tensor of size $28 \times 144$. It’s important to note that for missing teeth, their state values are set to zero.

\subsubsection{Action Space ($\mathcal{A}$)}
An action $a_t \in \mathcal{A}$ represents the movements of all teeth at timestep $t$. For each tooth $i$, the movement is described by a combination of a translation $\Delta p_{i,t} \in \mathbb{R}^3$ and a rotation $\Delta \theta_{i,t} \in \mathbb{R}^6$. The action $a_t \in \mathbb{R}^{28\times9}$ is then as follows:
\begin{equation}    
a_t = \{[\Delta p_{i,t}, \Delta \theta_{i,t}]\}_{i=1,2,...,28}
\end{equation}
All action values are normalized to the range $[-1, 1]$ to facilitate policy learning. Additionally, the actions predicted by the policy are scaled to lie within a predefined range, ensuring compliance with the physiological limits of clinical orthodontics.

\subsubsection{Transition Function ($\mathcal{P}$)}
Given the current state $s_t$ and action $a_t$, the transition function $\mathcal{P}(s_{t+1} | s_t, a_t)$ forms a new state $s_{t+1}$ by setting the last action to $a_t$ and calculating both the current pose and the current-to-target pose difference for each tooth $i$, as follows:
\begin{equation}
\label{eq:transition}
{pose}_{i,t+1} = {pose}_{i,t} + a_{i,t}
\end{equation}
\begin{equation}
\label{eq:transition}
{\Delta pose}_{i,t+1} = {pose}_{i,tgt} - {pose}_{i,t+1}
\end{equation}
% \textbf{where $+$ and $-$ denote element-wise addition and subtraction of the pose vectors, respectively.}

\subsubsection{Reward Function ($\mathcal{R}$)}
The reward function in an MDP provides crucial feedback to guide the agent toward optimal behavior. In 3D geometric tooth alignment planning, the goal is to generate a trajectory that effectively transitions teeth to the target alignment while satisfying spatial constraints—specifically avoiding inter-dental collisions and optimizing path efficiency. Based on these objectives, the reward $r_t$ at timestep $t$ is formulated to incorporate the following components:

\textbf{Progress Reward - $R_{prog}$:} 
This term rewards actions that bring the teeth closer to the target pose and penalizes those that move them farther away. It compares the distances between the current and target poses before and after an action using a fractional and logarithmic function. The reward is computed separately for translation and rotation as follows:
\begin{itemize}
    \item Translation Reward:
    \begin{equation}
    \label{eq:reward_trans}
    % R_{trans} = \mathbb{E}_{i} \left[ -\log\frac{1+k_t \mathcal{D}_p^2(p_{i,t}+\Delta{p_{i,t}})}{1+k_t \mathcal{D}_p^2(p_{i,t})} \right]
    % \end{equation}
    R_{trans} = \frac{1}{N} \sum_{i=1}^{N}  -\log\frac{1+k_t \mathcal{D}_p^2(p_{i,t}+\Delta{p_{i,t}})}{1+k_t \mathcal{D}_p^2(p_{i,t})}
    \end{equation}
    where $\mathcal{D}_p^2(p_{i,t})$ is the squared Euclidean distance between the position $p_{i,t}$ of tooth $i$ at time $t$ and its target $p_{i, tgt}$, $\Delta{p_{i,t}}$ is the translational component of the action, and $N$ is the number of teeth. The negative logarithm ensures a positive reward if the distance decreases, and a negative reward if it increases. The parameter $k_t$ scales the distance, and the ``+1'' ensures numerical stability. 
    % The expectation $\mathbb{E}_{i}$ averages over all 28 teeth indexed by $i$.
    \item Rotation Reward:
    \begin{equation}
    \label{eq:reward_rot}
    % R_{rot} = \mathbb{E}_{i} \left[ -\log\frac{1+k_r \mathcal{D}_r^2(r_{i,t}+\Delta{r_{i,t}})}{1+k_r \mathcal{D}_r^2(r_{i,t})} \right]
    R_{rot}= \frac{1}{N} \sum_{i=1}^{N}  -\log\frac{1+k_\theta \mathcal{D}_\theta^2(\theta_{i,t}+\Delta{\theta_{i,t}})}{1+k_\theta \mathcal{D}_\theta^2(\theta_{i,t})}
    \end{equation}
    This is formulated similarly to the translation reward, except $\mathcal{D}_\theta^2(\theta_{i,t})$ computes the squared distance in angular space.
\end{itemize}
The total progress reward is:
\begin{equation}
\label{eq:reward_prog}
R_{prog}=\alpha_{trans}R_{trans}+\alpha_{rot}R_{rot}
\end{equation}
where $\alpha_{trans}$ and $\alpha_{rot}$ are weighting factors for translation and rotation rewards.

\textbf{Collision Penalty - $R_{coll}$:} 
This term penalizes unsafe tooth movements that lead to abnormal inter-tooth collisions. We first use the well-known GJK algorithm \cite{gilbert2002fast} to detect collisions and calculate the penetration depth between two teeth to assess the severity of the collision. Next, we design a piecewise linear function that applies varying negative rewards based on the penetration depth. The function is as follows:
\begin{equation}
\label{eq:penalty_func_final}
P(\delta) = 
\begin{cases}
    -60, & \text{if } \delta > 0.3\,\text{mm}; \\
    -40, & \text{if } 0.25\,\text{mm} < \delta \le 0.3\,\text{mm}; \\
    -20, & \text{if } 0.2\,\text{mm} < \delta \le 0.25\,\text{mm}; \\
    -5,  & \text{if } 0.15\,\text{mm} < \delta \le 0.2\,\text{mm}; \\
    0,   & \text{otherwise.}
\end{cases}
\end{equation}
where $\delta$ represents the penetration depth. The final penalty is the sum over all $J$ potential colliding pairs (a pair is defined as two adjacent teeth in this study): $R_{coll} = \sum_{j}^{J} P(\delta_j)$.

\textbf{Smoothness Penalty - $R_{smooth}$:} 
To ensure a smooth path, we impose a negative reward for sudden changes between two consecutive orthodontic steps. Specifically, we use the L1 distance between the translational components of two consecutive actions to quantify the change. The reward is represented as the negative average distance across all teeth, as follows:
\begin{equation}
\label{eq:reward_smooth}
R_{smooth} = -\frac{\alpha_{smooth}}{N} \sum_{i=1}^{N} || \Delta p_{i,t} - \Delta p_{i,t-1} ||_1
\end{equation}
where $N$ is the number of teeth and $\alpha_{smooth}$ is the hyperparameter that weights the reward term.

\textbf{Terminal Bonus - $R_{terminal}$:} 
A terminal bonus is awarded if the new state achieves the goal, which is when all teeth are within 0.2 mm of their target positions and 3 degrees of their target orientations. The bonus is then defined as:
\begin{equation}
\label{eq:terminal_bonus}
\hspace*{-3pt} R_{terminal}(s_t, a_t) = 
\begin{cases}
    100, & \text{if } s_t+a_t \text{ is a goal state;} \\
    0, & \text{otherwise.}
\end{cases}
\end{equation}

\textbf{Total Reward:} The final reward $R(s_t, a_t)$ for transitioning from the current state $s_t$ to a new state via action $a_t$ is defined as:
\begin{equation}
\label{eq:total_reward}
\mathcal{R}(s_t, a_t) = R_{prog} + R_{coll} + R_{smooth} + R_{terminal}
\end{equation}

\subsection{Adapting DDPG to Solve the MDP}    
We use the well-established Deep Deterministic Policy Gradient (DDPG) algorithm \cite{lillicrap2015continuous} to solve the MDP defined above. DDPG is an off-policy reinforcement learning method that employs deep neural networks to approximate both the policy (actor) and the value function (critic) in continuous action spaces. It combines the benefits of deterministic policy gradients with the stability provided by experience replay and target networks. DDPG involves four networks: the actor, the target actor, the critic, and the target critic. These networks form the core of the agent. The target networks maintain slowly updated copies of the actor and critic networks, which help stabilize training by smoothing updates to the value function and policy. (For a detailed explanation of DDPG, please refer to \cite{lillicrap2015continuous}).

Despite DDPG's effectiveness, the high-dimensional state ($s_t \in \mathbb{R}^{28\times144}$) and action ($a_t \in \mathbb{R}^{28\times9}$) spaces, along with the structural complexity of real-world dentition, present significant challenges for policy learning. To address these challenges, we propose three key adaptations to tailor the vanilla DDPG framework to the 3D geometric tooth alignment planning MDP. These adaptations include: a transformer-based agent, a dynamic masking scheme to enforce action sparsity, and a two-stage curriculum learning strategy, which are detailed as follows:

\subsubsection{Transformer-based Agent with Dynamic Action Masking}
The actor and critic networks form the core of the agent in DDPG. These networks are implemented using a transformer backbone \cite{vaswani2017attention}, with multi-head attention applied along the tooth dimension of the input tensor to capture inter-tooth relationships. Tooth location (with upper and lower jaws handled independently) is encoded using absolute positional encoding based on each tooth's index number (from 0 to 13, ordered left to right within the arrangement). This positional encoding is then added to the input tensor. To enrich the positional information further, we introduce a learnable relative positional encoding scheme \cite{press2021train}. This scheme adds a learnable bias term $B$ to the standard attention score calculation:
\begin{equation}
\label{eq:attention}
\text{Attention}(Q, K, V) = \text{softmax}\left(\frac{QK^T}{\sqrt{d_k}} + B\right)V
\end{equation}
where the bias $B_{i,j} = -\mu \cdot \rho_{ij}$ is a function of the index distance $\rho_{ij}$ between teeth $i$ and $j$, scaled by a learnable parameter $\mu$. This mechanism forces the model to produce a higher attention score between two teeth that are spatially closer.

\textbf{Actor Network.} The actor network $\pi(\cdot)$ should learn not only \textit{how} to move each tooth from the state tensor $s_t$, but also \textit{which} teeth to move. To achieve this, we attach two predictive heads to the transformer backbone: an action head and a mask head. The action head generates a base action tensor $a_{base,t} \in \mathbb{R}^{28\times9}$, while the mask head generates a mask vector $mask_{a,t} \in [0, 1]^{28 \times 1}$, where each element corresponds to a specific tooth and its value indicates the probability that the tooth should be moved at timestep $t$. The final action is the result of the element-wise product: $a_t = a_{base,t} \odot mask_{a,t}$. In clinical practice, sequential tooth movements exhibit significant sparsity, where only a sparse subset of teeth is adjusted at each step. To encourage the agent to mimic this behavior within our 3D geometric tooth alignment planning framework, we impose a sparsity constraint on the predicted action mask. Specifically, we employ a regularization term that drives the mask values towards a binary distribution and penalizes an excessive number of active movements. The formulation is as follows:
\begin{align}
\label{eq:sparsity_loss}
\mathcal{L}_{sparsity} &= \alpha_{sum} \cdot \max\left(\text{sum}(mask_{a,t}) - \tau, 0 \right) \\
&\quad + \alpha_{binary} \cdot (mask_{a,t} \cdot (1 - mask_{a,t})) \notag
\end{align}
where $\alpha_{sum}$ and $\alpha_{binary}$ are weighting coefficients, and $\tau$ is a manually set threshold that defines the starting point for unnecessary tooth movements. 
The actor network is trained to maximize the Q-value output by the critic network $Q(\cdot)$, with regularization provided by the sparsity term mentioned above. The total loss is:
\begin{equation}
\label{eq:actor_loss}
\hspace*{-2pt} \mathcal{L}_{\pi} = -\mathbb{E}_{s \sim \mathcal{S}} [Q(s, \pi(s).a) - \mathcal{L}_{sparsity}(\pi(s).mask)] 
% \mathcal{L}_{\pi} = -\mathbb{E}_{s \sim \mathcal{S}} [Q(s, \pi(s).a)] + \mathcal{L}_{sparsity}(\pi(s).mask)
\end{equation}
where $\pi(s).a$ and $\pi(s).mask$ represent the action and mask predicted by the actor $\pi(\cdot)$ given the state $s$.

\textbf{Critic Network.} The critic network $Q(\cdot)$ takes the state $s_t$ and action $a_t$ as input and outputs a Q-value representing the expected cumulative reward for taking action $a_t$ in state $s_t$. It is trained to minimize the Mean Squared Error (MSE) between its predicted Q-value and a target $y_t$, which is based on the rewards received and the estimated future rewards (from the next states). To balance the low bias of multi-step returns with the low variance of the single-step return, we construct the target $y_t$ as a soft mixture of the 1-step and N-step returns:
\begin{equation}
\label{eq:td_target_combined}
y_t = 0.5 \cdot y_t^{(1)} + 0.5 \cdot y_t^{(N)}
\end{equation}
where $y_t^{(1)}$ is the 1-step target and $y_t^{(N)}$ is the N-step target, defined as follows:
\begin{align}    
y_t^{(1)} &= r_t + \gamma Q_{\text{tgt}}(s_{t+1}, \pi_{\text{tgt}}(s_{t+1})) \label{eq:1_step_target} \\    
y_t^{(N)} &= \sum_{k=0}^{N-1} \gamma^k r_{t+k} + \gamma^N Q_{\text{tgt}}(s_{t+N}, \pi_{\text{tgt}}(s_{t+N})) \label{eq:N_step_target}
\end{align}
Here, $Q_{\text{tgt}}$ and $\pi_{\text{tgt}}$ represent the target networks, which are time-delayed copies of the main networks used to stabilize training by preventing the critic's Q-value estimates from changing too quickly. $\gamma$ is the discount factor that determines the weight given to future rewards compared to immediate rewards. The final critic loss is then computed over batches sampled from the replay buffer $\mathcal{D}$:
\begin{equation}
\label{eq:critic_loss_final}
\mathcal{L}_{Q} = \mathbb{E}_{(s_t,a_t,r_t,\dots) \sim \mathcal{D}} \left[ (Q(s_t,a_t) - y_t)^2 \right]
\end{equation}

\subsubsection{Training Regimen with Two-Stage Curriculum Learning}
 To train our agent in the high-dimensional action space, we design a two-stage curriculum learning \cite{bengio2009curriculum} approach, enhanced with established reinforcement learning techniques to improve efficiency. The approach primarily uses a coarse-to-fine curriculum to address the problem's complexity.
 
% \textbf{Stage 1: Coarse Exploration.} The goal of the first stage is to quickly identify a general path to the target. This is achieved by using high learning rates for both the critic ($\eta_q = 10^{-3}$) and actor ($\eta_{\pi} = 10^{-4}$), both of which are linearly annealed. Penalties for collision and smoothness are relaxed to prioritize goal achievement. A gentle penalty of $R_{coll}=10$ is applied for severe collisions with $\delta > 0.3$\,mm, while the smoothness penalty weight is set to $\alpha_{smooth}=10$. Additionally, the loss term on the action mask is set (with $\alpha_{binary}=0$ and $\tau=6$) to encourage moving more teeth at each step, accelerating progress.

\textbf{Stage 1: Coarse Exploration.} 
The goal of the first stage is to quickly identify a general path to the target. This is achieved by using high learning rates for both the critic ($\eta_q = 10^{-3}$) and actor ($\eta_{\pi} = 10^{-4}$), both of which are linearly annealed. Penalties for collision and smoothness are relaxed to prioritize goal achievement.
Specifically, instead of the piecewise penalty defined in Eq.~(7), we adopt a simplified binary collision penalty: a fixed penalty of $R_{coll}=-10$ is imposed only when a severe collision occurs ($\delta > 0.3$\,mm); otherwise, $R_{coll}=0$. This relaxation weakens the constraint while still discouraging unsafe behaviors, thereby promoting exploration of coarse trajectories toward the goal. The smoothness penalty weight is set to $\alpha_{smooth}=10$. Additionally, the loss term on the action mask is set (with $\alpha_{binary}=0$ and $\tau=6$) to encourage moving more teeth at each step, accelerating progress.

\textbf{Stage 2: Path Refinement.} This stage focuses on refining the path for clinical safety and efficiency. The learning rate is reduced for stable convergence (critic - $\eta_q = 10^{-4}$, actor - $\eta_{\pi} = 10^{-5}$), and both are linearly annealed as in Stage~1. Penalties for collision and smoothness are significantly increased (the collision penalty follows~\eqref{eq:penalty_func_final}, and $\alpha_{smooth}=200$) to enforce stricter constraints. The loss on the action mask ($\alpha_{binary}=100$, $\tau=4$) is increased to encourage tooth movements that are more clinically realistic.
Across both stages, we employ three key techniques to enhance training efficiency:
\begin{itemize}    
\item N-Step Learning \cite{sutton1998reinforcement}: We use N-step returns ($N=3$) to accelerate reward propagation and improve credit assignment.    
\item Prioritized Experience Replay (PER) \cite{schaul2015prioritized}: A PER buffer is used to focus training on informative transitions by replaying them more frequently.    
\item Learning from Demonstrations \cite{hester2018deep,schaal1996learning,vecerik2017leveraging}: The replay buffer is pre-populated with expert trajectories, which are permanently retained and prioritized. This bootstraps the learning process and provides continuous guidance toward clinically plausible solutions.
\end{itemize}

\section{Experiments}
\label{sec:experiments}
We thoroughly evaluate the proposed method by comparing it with existing methods and conducting an in-depth ablation study. 

\subsection{Implementation Details}
We tune key hyperparameters individually. Most hyperparameters correspond to reward terms, which are added incrementally - starting with $R_{terminal}$, then progressively including $R_{coll}$, $R_{trans}$, $R_{rot}$, and $R_{smooth}$. Each newly introduced reward term is assigned an initial weight and subsequently adjusted to maintain comparable reward magnitudes and stable training dynamics. The adjustment is guided by monitoring reward scale, gradient stability, and convergence behavior, rather than exhaustive hyperparameter search.  Some hyperparameters are determined based on training data statistics. For instance, the threshold $\tau$ in the dynamic masking mechanism constrains the number of moving teeth per step. Data analysis shows an average of 5.39 moving teeth per step in expert trajectories; therefore, $\tau=6$ is used in Stage 1 to match this pattern and encourage exploration, while $\tau=4$ in Stage 2 produces results more consistent with expert data.

All experiments are conducted on a single NVIDIA RTX 4090 GPU. The training process comprises 4 million steps and takes approximately 32 hours to complete. During inference, generating a complete alignment trajectory for a single case requires approximately 0.3–0.42 seconds.

\subsection{Dataset}
Our methodology is evaluated on a large-scale dataset of 10K expert-designed orthodontic plans based on clinical data, provided by Choho Technology, a startup dedicated to advancing intelligent oral healthcare. Each case includes 3D tooth meshes and the corresponding ground-truth tooth movement path designed by doctors. Notably, the dataset assumes constant tooth geometry and omits any shape alterations during treatment, such as those from interproximal reduction. While this idealization minorly departs from clinical reality, where dynamic reshaping is often required to resolve inter-tooth collisions, it maintains computational feasibility. The dataset is split into a training set of 9K cases, which serve as expert demonstrations to populate the prioritized replay buffer in our method, and a held-out test set of 1K cases for final performance evaluation.

\subsection{Evaluation Metrics}
We use a series of metrics to evaluate the efficiency and safety of the tooth movement path produced by different methods:
\begin{itemize}    
\item $sum_T$: The total cumulative translation of the tooth path, measured in millimeters. A lower value indicates higher path efficiency.    
\item $sum_R$: The total cumulative rotation of the tooth path, measured in radians. A lower value is preferred for higher efficiency.    
\item $N_{violate}$: The number of steps in which any tooth exceeds the clinical movement limit. We follow \cite{martinez2024staging}, where the translation limit is 0.25mm and the rotation limit is 3$^\circ$.    
\item $f_{collision}$: The average frequency of collisions, calculated as the collision count divided by the tooth count. Collisions are detected by extracting the convex hull of each tooth and using the Gilbert-Johnson-Keerthi (GJK) algorithm \cite{gilbert2002fast}. A penetration depth threshold of 0.3 mm is used to define collisions.    
\item $\Delta{N}$: The absolute difference in the number of steps between the generated path and the corresponding expert-designed path. This metric is only applicable to methods that can generate paths of varying lengths, and provides an indication of the resemblance between the predicted result and the expert data.
\end{itemize}

\subsection{Results}
We compare our method with three strong baselines: 1) IGWO \cite{du2023orthodontic}, an optimization-based approach that employs the Improved Gray Wolf Optimizer to solve tooth motion planning; 2) NeuralOrtho \cite{ma2024neural}, a learning-based method that generates the tooth movement path iteratively, using a transformer to predict a fixed-length sequence at each stage; and 3) TMDM \cite{fan2024collaborative}, an approach that utilizes a diffusion model to generate the tooth motion path over a predefined length. For a fair comparison, both IGWO~\cite{du2023orthodontic} and TMDM~\cite{fan2024collaborative} baselines follow the same experimental assumptions as in their original papers, using the same path lengths as the expert data.

\begin{table}
\caption{Quantitative Comparison of Our Method with Baseline Approaches on the Test Set}
\label{tab:comparisons}
\begin{tabular*}{\linewidth}{@{\extracolsep{\fill}}lccccc}
\hline
\textbf{Method} & $sum_{T}$$\downarrow$  & $sum_{R}$$\downarrow$ & $N_{violate}$$\downarrow$ & $f_{collision}$$\downarrow$ & $\Delta{N}$$\downarrow$ \\ 
& (mm) & (rad) &  &  &  \\
\hline
Expert & 57.72 & 6.72 & 2.77 & 0.194 & $-$ \\
\hline
IGWO & 68.72 & 8.67  & 1.12 & 0.331 & $-$ \\
NeuralOrtho & \textbf{56.12} & \textbf{5.93} & 1.62 & 0.267 & 13.49\\
TMDM & 63.93 & 49.87 & 331.90 & 0.249 & $-$ \\
\hline
\textbf{Ours} & 59.91 & 7.01 & \textbf{0.52} & \textbf{0.237} & \textbf{9.77} \\
\hline
\end{tabular*}
\end{table}

The quantitative comparison results are summarized in Table~\ref{tab:comparisons}. Our method outperforms all baselines in safety metrics, producing paths with much fewer violations ($N_{violate}$) and collisions ($f_{collision}$). It achieves second place in path efficiency, as measured by $sum_T$ and $sum_R$. Notably, although the NeuralOrtho method shows slightly lower total translation and rotation, this efficiency comes at a significant cost: it exhibits a much higher collision rate and more constraint violations. Additionally, our method shows the smallest difference in path length ($\Delta{N}$) when compared to the expert data. In contrast, a key structural limitation of both IGWO and TMDM is their requirement to predefine the length of the tooth movement path, which does not align with the varying-length nature of 3D geometric tooth alignment planning.

\begin{figure}[t]       
\centerline{\includegraphics[width=\columnwidth]{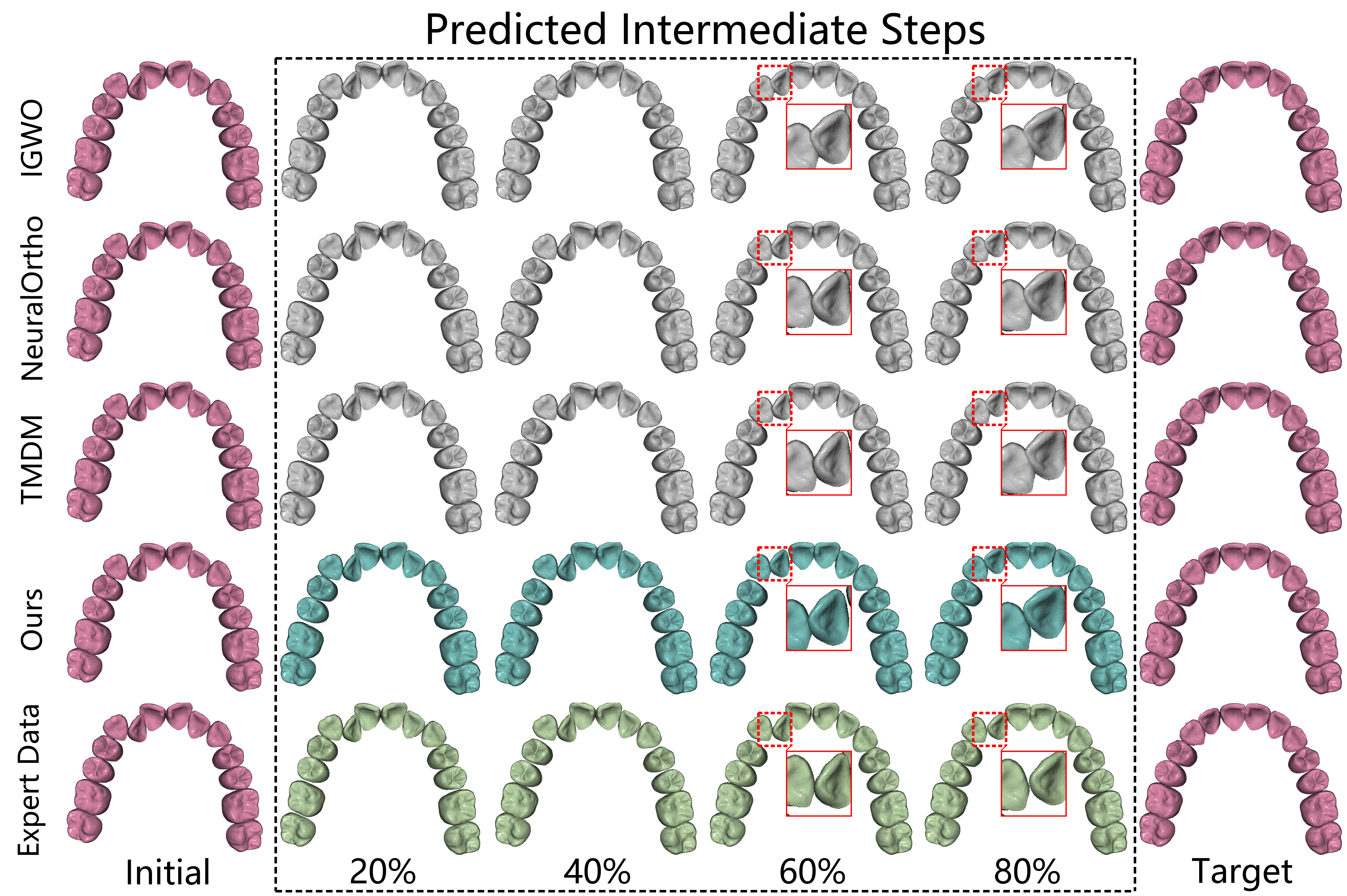}}
\caption{Visual comparison of our method with three state-of-the-art methods. As highlighted by the red boxes, our method generates a tooth movement path with fewer collisions and that closely resembles the expert path.}    
\label{fig:collision_comparison}
\end{figure}

\begin{figure*}[t]
    \includegraphics[width=0.95\textwidth]{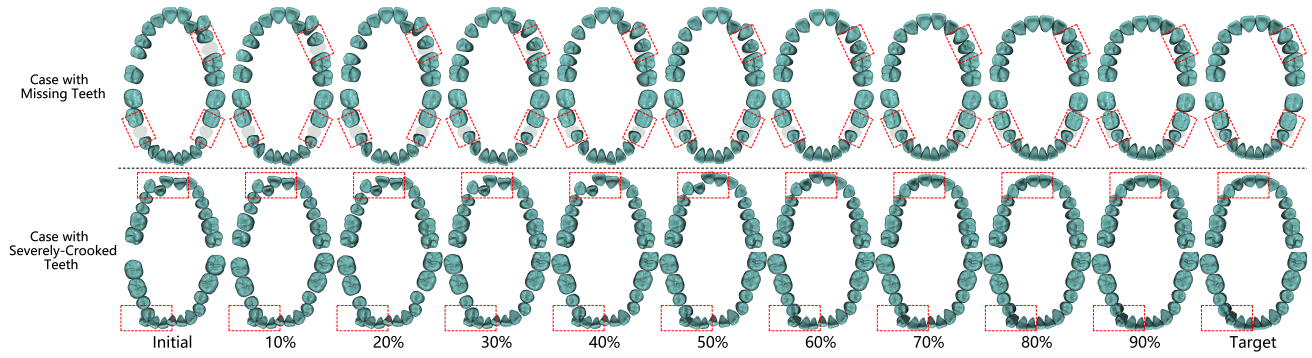}
    \caption{The visual results of our method on two challenging cases: one with missing teeth and another with severely crooked teeth. Our method effectively addresses the dental complexities (highlighted by red dashed boxes) in these cases, generating a smooth and collision-free path. (The percentage represents the progress of the path.)}
    \vspace{-5pt}
    \label{fig:challenging_cases}
\end{figure*}

We also provide a visual comparison of the methods using a challenging case involving severe tooth misalignment, shown in Fig.~\ref{fig:collision_comparison}. The results demonstrate that our method avoids unreasonable collisions in the generated path, while the three baseline methods exhibit clear problematic collisions during the intermediate tooth movement steps. Additional visual results of our method are shown in Fig.~\ref{fig:challenging_cases}. As illustrated, our method effectively handles dental intricacies such as absent teeth and extremely crooked teeth. For instance, in the lower case, our method first moves the molars to create space for the crowded and severely twisted incisors. This highlights our method's ability to produce physically plausible and geometrically sound alignment trajectories. \textbf{Additional results are provided in the supplementary material.}

\begin{figure}[h]    
\centering    
\includegraphics[width=0.90\linewidth]{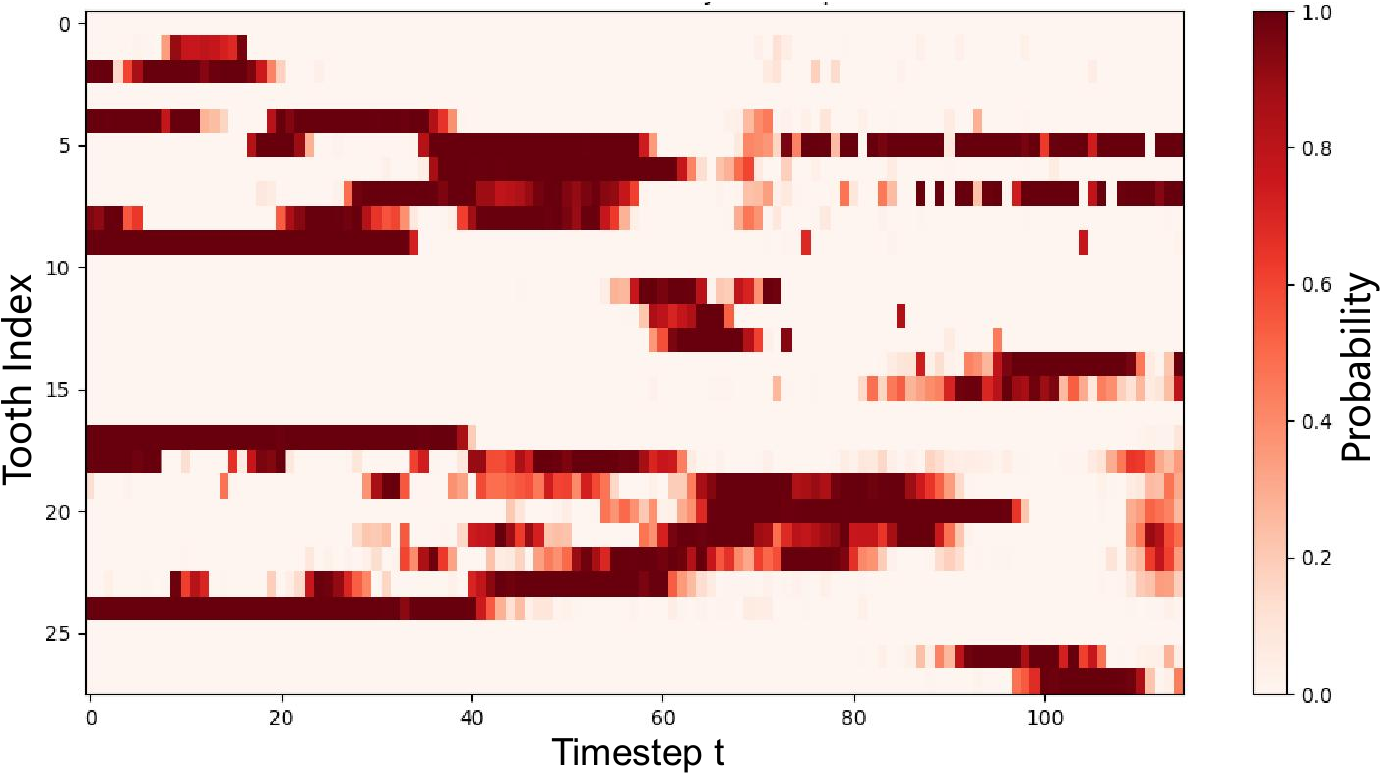}    
\caption{Action Mask Heatmap. Horizontal axis: timestep $t$. Vertical axis: tooth index. Each cell in the heatmap indicates the probability that the corresponding tooth is being moved at that specific timestep. The sparse and temporally coherent activation patterns align with clinical staging behavior.}    
\label{fig:mask_heatmap}
\end{figure}

To illustrate the action sparsity induced by our dynamic action-masking mechanism, we visualize the predicted action masks over an entire orthodontic trajectory. At each timestep \(t\), the actor outputs a mask vector \(\mathrm{mask}_{a,t} \in [0,1]^{28 \times 1}\), where each dimension corresponds to a specific tooth index. Fig.~\ref{fig:mask_heatmap} shows a heatmap of these mask values, with timesteps on the horizontal axis and tooth indices (0--27) on the vertical axis. Each cell represents the predicted probability that a particular tooth is selected for movement at that timestep. The visualization clearly shows that only a sparse subset of teeth is activated at each step, consistent with clinical orthodontic practice in which only a limited number of teeth are moved simultaneously. Moreover, the activated tooth groups form short, temporally coherent segments, reflecting stable and biologically plausible movement patterns. This further confirms that the dynamic masking mechanism effectively constrains the action space and guides the policy toward clinically realistic behaviors.

\subsection{Ablation Study}
We conduct an in-depth ablation study to validate the effectiveness of our key algorithmic designs, including the transformer-based agent and the two-stage curriculum learning strategy.

\textit{Validation of Transformer-based Agent.} We ablate the core components of our transformer agent to isolate their individual contributions, with results presented in Table~\ref{tab:ablation_components}. Removing either expert data or the action masking mechanism leads to a catastrophic performance collapse (e.g., 0\% and 70.5\% success rates (SR), where SR is defined as the percentage of cases in which the agent reaches the target pose within 200 steps. The upper limit of 200 steps is chosen based on statistical analysis, since all training cases have path lengths below 150.), confirming that both are foundational to our framework. Ablating other components, including relative positional encoding, N-Step learning, and prioritized experience replay (PER), also results in noticeable degradation in path quality, highlighting their critical role in refining the policy. Together, these results demonstrate that our full model achieves the best synergy for producing safe and efficient tooth movement paths.

\begin{table}[t]
\caption{Ablation Study on the Core Components of the Transformer-based Agent}
\label{tab:ablation_components}
\begin{tabular*}{\linewidth}{@{\extracolsep{\fill}}lccccc}
\hline
\textbf{Method} & SR$\uparrow$ & $sum_{T}$$\downarrow$ & $sum_R$$\downarrow$ & \textbf{$f_{collision}$}$\downarrow$ & $\Delta N$$\downarrow$ \\
 & (\%) & (mm) & (rad) &  &  \\
\hline
\textbf{Ours (Full)} & \textbf{99.9} & \textbf{59.91} & \textbf{7.01}  & \textbf{0.237} & \textbf{9.8}  \\
\hline
% w/o (use MLP) & 70.5 & 115.06 & 17.77 & 58.1 & 381 & 0.948 \\
w/o Act. Mask. & 70.5 & 115.06 & 17.77 & 0.948 & 58.1 \\
\hline
\multicolumn{6}{l}{\textit{Components ablated from Stage 2}} \\
w/o Rel. Pos. Enc. & 99.8 & 61.17 & 7.64 & 0.269 & 10.0 \\
w/o N-Step Learn. & 95.9 & 70.35 & 8.35  & 0.394 & 17.4 \\
w/o Expert Data & 0 & 259.74 & 31.31  & 3.487 & 152.3 \\
w/o PER & 99.3 & 67.18 & 7.67 & 0.248  & 10.8  \\
\hline
\end{tabular*}
\end{table}

\textit{Efficacy of Two-Stage Curriculum Learning.} We compare our two-stage training strategy with the vanilla single-stage method. The single-stage baseline is trained for a total of 4M steps, while our two-stage method is trained for 2M steps in Stage 1 and 2M steps in Stage 2. The switching phase at 2M steps is an empirical choice, as we observe the rewards plateau beyond this point. As shown in Table~\ref{tab:ablation_curriculum}, the single-stage approach completely fails to learn when using a low discount factor ($\gamma=0.90$), achieving a 0\% success rate (SR).
A lower $\gamma$ means the agent prioritizes immediate rewards, leading to a more exploitative strategy. Even with higher $\gamma$ values (which encourage the agent to prioritize future rewards and explore more), the single-stage baseline remains significantly inferior across most path quality metrics. This highlights the importance of the initial coarse exploration stage, which helps the agent discover a viable policy that is then refined in the second stage.

\begin{table}[t]
\caption{Comparison of the Two-Stage Curriculum Learning Strategy with the One-Stage Learning Strategy}
\label{tab:ablation_curriculum}
\begin{tabular*}{\linewidth}{@{\extracolsep{\fill}}lcccccc}
\hline
$\gamma$ & \textbf{Method} & SR$\uparrow$ & $sum_{T}$$\downarrow$ & $sum_{R}$$\downarrow$ & $f_{collision}$$\downarrow$ & $\Delta{N}$$\downarrow$\\
 &  & (\%) & (mm) & (rad) &  &  \\
\hline
0.90 & Single-Stage & 0 & \textbf{48.54} & 13.88 & 0.584 & 152.3\\
& \textbf{Two-Stage} & \textbf{99.3} & 61.51 & \textbf{7.59} & \textbf{0.236} & \textbf{10.9} \\
\hline
0.95 & Single-Stage & 98.6 & 68.34 & 7.75 & 0.264 & 11.4 \\
& \textbf{Two-Stage} & \textbf{99.9} & \textbf{59.91} & \textbf{7.01}  & \textbf{0.237} & \textbf{9.8} \\
\hline
0.98 & Single-Stage & 98.8 & 70.58 & 7.96 & 0.269  & 10.8 \\
& \textbf{Two-Stage} & \textbf{99.6} & \textbf{62.91} & \textbf{7.73} & \textbf{0.245}  & \textbf{10.1} \\
\hline
\end{tabular*}
\end{table}

\section{Limitations and Future Work}
\label{sec:limitations}
While this study validates the feasibility of reinforcement learning (RL) for automated 3D geometric tooth alignment planning, several limitations remain to be addressed in future work. First, our framework primarily aims to establish the viability of RL for this task rather than to provide a comprehensive comparison across modern RL algorithms. More advanced RL methods such as TD3~\cite{fujimoto2018addressing} and SAC~\cite{haarnoja2018soft} are not included in this study, though they may offer potential gains in performance and training stability. Second, as this represents the first attempt to introduce RL into this alignment planning task, the proposed framework involves a relatively large number of hyperparameters. While the current configuration yields stable and geometrically plausible results, there remains substantial room for improvement through systematic hyperparameter optimization to enhance robustness and generalization. Third, while actual orthodontic treatment is a highly complex process involving various biomechanical factors such as anchorage control and periodontal constraints, this work focuses on the geometric aspects of trajectory planning. Specifically, we explicitly model progress toward target alignment, collision avoidance, smoothness, and sparsity of tooth movements. However, certain clinical priors, such as explicit inter-dental gap regularization during intermediate stages, are not directly incorporated. Although final alignment implicitly enforces proper contact relationships, integrating additional clinical constraints may further improve intermediate trajectory realism. Exploring such structured priors within the RL framework is an important direction for future research.

\section{Conclusion}
\label{sec:conclusion}
In this study, we present a novel deep reinforcement learning-based approach to automated 3D geometric tooth alignment planning. The method frames the problem as a Markov Decision Process (MDP) to capture the sequential decision-making nature of the task. Based on the Deep Deterministic Policy Gradient (DDPG) algorithm \cite{lillicrap2015continuous}, we propose a new framework for addressing the alignment planning MDP. The framework features a transformer-based agent, enhanced with a dynamic action masking mechanism for sparse action control and a two-stage curriculum learning strategy for effective model training. Extensive experiments on a dataset of 10K expert-designed orthodontic pathways based on clinical data demonstrate that our method outperforms existing approaches, generating collision-free and more efficient tooth movement trajectories to establish a new state-of-the-art in the field.

% \section*{REFERENCES}

\bibliographystyle{IEEEtran}
\bibliography{ref}

\end{document}